\pdfoutput=1
\documentclass[11pt,a4paper]{article}
\usepackage[scaled=0.85]{helvet}
\usepackage{times}
\usepackage[final]{acl}
\usepackage{latexsym}
\usepackage[T1]{fontenc}
\usepackage[utf8]{inputenc}
\usepackage{lineno}
\usepackage{microtype}
\usepackage{inconsolata}
\usepackage{calc}
\usepackage{graphicx}
\usepackage{url}
\usepackage{enumitem}
\usepackage{dirtytalk}
\usepackage{amssymb}
\usepackage{booktabs}
\usepackage{multirow}
\usepackage{makecell}
\usepackage{adjustbox}
\usepackage{amsmath}
\usepackage{tabularx}
\usepackage{tikz}
\usetikzlibrary{arrows.meta,shapes.geometric,positioning}

\usepackage[main=american]{babel}

\useshorthands*{"}
\defineshorthand{"-}{\babelhyphen{hard}}
\defineshorthand{"=}{\babelhyphen{–}}
\defineshorthand{"/}{\babelhyphen{/}}

\newcommand*{\CORPUS}[1]{\textit{#1}}
\newcommand*{\SETUP}[1]{\textbf{#1}}

\title{Outgrowing the Curse of Multilinguality:\\ Scaling Strategies for Efficient Language Adaptation}

\author{  Kevin Glocker$^*$ \quad Kätriin Kukk$^*$ \quad Romina Oji \\ \textbf{Marcel Bollmann} \quad 
  \textbf{Marco Kuhlmann} \quad \textbf{Jenny Kunz} \\[\smallskipamount]
  Department of Computer and Information Science \\
  Linköping University \\
  \texttt{firstname.lastname@liu.se} \\}

\begin{document}
\maketitle
\begin{abstract}
Training high"-performing language models which include medium"- and lower"-resource languages remains a challenge. Massively multilingual models still underperform language"-specific adaptations, especially at smaller scales, due to the \emph{curse of multilinguality}. In this work, we investigate \emph{scaling} as an efficient strategy for adapting pretrained models to new target languages. Through comprehensive ablations matched for data and training FLOPs, we test whether upscaling from English base models enables more effective and resource"-efficient adaptation than standard continued pretraining. We find that upscaling is an efficient strategy for improving target-language performance by relieving capacity constraints, especially for higher"-resourced languages. An initial upscaling step in English is beneficial for low-resource languages. However, in higher-resource scenarios it is more compute-efficient to directly scale up on the target language, albeit slightly compromising data efficiency for most languages.
For multilingual continued pre-training, we identify directly scaling up using the target data as the optimal setup for both data efficiency and reduced English forgetting.

\end{abstract}

\section{Introduction}

\def\thefootnote{*}\footnotetext{Equal contribution.}
\renewcommand{\thefootnote}{\arabic{footnote}}

Massively multilingual language models \citep{mesnard-etal-2024-gemma, ji2025emma500enhancingmassivelymultilingual} are widely used in real"-world applications, yet their performance remains uneven across languages.
Especially at smaller scales, they still face the \emph{curse of multilinguality} \citep{conneau2019_roberta}. The need to share limited capacity across many languages often results in low performance, particularly for low"-resource languages.
Indeed, multilingual models can be outperformed by much smaller monolingual counterparts \citep{chang2024goldfishmonolinguallanguagemodels}.
A common remedy for this performance gap is continued pretraining on target"-language data, often using parameter"-efficient finetuning methods \citep{pfeiffer-etal-2020-mad, razumovskaia-lora}.
However, evidence suggests that for smaller models, full"-parameter finetuning yields better results than parameter"-efficient alternatives \citep{yong-etal-2023-bloom}.

\begin{figure}
    \centering
    \definecolor{basefill}{HTML}{EBCB8B}
\definecolor{targetfill}{HTML}{A3BE8C}
\definecolor{upscalefill}{HTML}{5E81AC}

\begin{tikzpicture}[%
    yscale=0.98,xscale=0.98,every node/.style={scale=0.8},
    node distance=1.2cm,
    model/.style={rectangle, rounded corners, thick, draw=black, align=center, fill=basefill!50, inner sep=2pt},
    target/.style={fill=targetfill!50},
    foundation/.style={fill=upscalefill!15},
    data/.style={rectangle, fill=white, font={\small}, align=center},
    modelname/.style={font={\bfseries}},
    upscale/.style={diamond, draw, thick, fill=upscalefill!15, font={\small}, inner sep=1pt},
    1x/.style={minimum width=1cm, minimum height=1cm},
    2x/.style={minimum width=2cm, minimum height=1cm},
    X/.tip={Triangle[]},
    arrow/.style={thick,rounded corners},
    ]

    \node[model,1x,foundation] (en80)      at ( 0,      0)  {180M};
    \node[model,1x] (en_1x)     at (-1.75, 1.5)  {180M};
    \node[model,1x,target] (target_1x) at (-1.75, 3.5)  {180M};

    \node[model,2x] (en_2x)       at (2, 2.5) {572M};
    \node[model,2x,target] (target_1x2x) at (0, 4.25) {572M};
    \node[model,2x,target] (target_2x)   at (2, 5.0) {572M};

    \node[modelname,yshift=-.4cm,above of=target_1x]   {1x};
    \node[modelname,yshift=-.4cm,above of=target_1x2x] {1x cloned};
    \node[modelname,yshift=-.4cm,above of=target_2x]   {2x};

    \draw[-X,arrow] (en80)   -| (en_1x);
    \draw[-X,arrow] (en_1x)  -- (target_1x);
    \draw[-X,arrow] (en80)   -| (en_2x);
    \draw[-X,arrow] (en_1x)  -| (target_1x2x);
    \draw[-X,arrow] (en_2x)  -- (target_2x);

    \node[data] (en80_data) at (0, -.8)  {\texttt{eng} (80\%)};
    \node[data,below of=en_1x] {\texttt{eng} (20\%)};
    \node[data,below of=target_1x] {Target};
    \node[data,below of=target_2x] {Target};
    \node[data,below of=en_2x] {\texttt{eng} (20\%)};
    \node[data,below of=target_1x2x] {Target};

    \node[right of=en80, upscale] (up1) {2$\times$};
    \node[right of=en_1x, upscale] (up2) {2$\times$};

    \draw[-X,ultra thick] (-3, -.5) -- (-3, 5.5) node [midway, rotate=90, fill=white] {Compute cost};
\end{tikzpicture}
    \caption{Illustration of our scaling setups.  Base models are in yellow, target-language models in green. Arrows represent continued pretraining, with ``2×'' indicating where we use hypercloning to upscale models.}
    \label{fig:scaling-setups}
\end{figure}

In this work, we therefore explore \emph{scaling} for adapting pretrained models to new languages.
Prior research shows that \emph{upcycling} smaller models can be substantially more efficient than training large models from scratch while achieving comparable or better downstream performance.
This is variously referred to as model growth \citep{du2024stacking}, model expansion \citep{pmlr-v262-samragh24a}, or scaling \citep{wang2025tokenformer}.
Although these methods have proven effective in monolingual contexts, their potential for language adaptation remains underexplored.
In this work, we extend this line of inquiry to the multilingual setting.
Specifically, we employ HyperCloning \citep{pmlr-v262-samragh24a}, which increases the dimensionality of the model's hidden layers.
Crucially, HyperCloning preserves the smaller model's output distribution, thereby giving the larger model a ``warm start'' by retaining its original capabilities before continued training.
Through an extensive set of experiments, we investigate whether scaling can improve language adaptation, and study the trade"-offs between compute and data requirements across scaling setups.

Our results show that upscaling is an effective strategy for language adaptation, producing models that are more data"-efficient and perform better than non"-scaled models.
We also show that for lesser"-resourced target languages, it is beneficial to first continue training in English after upscaling. In contrast, upscaling directly in the target language is more compute-efficient for higher-resourced languages. While this comes at a cost of data efficiency in the single-target-language setup, multilingually upscaled models are equally data-efficient even without an initial scaling step in English.

\section{Background and Related Work}

We describe the technical approach underlying our study, \textit{upscaling} (§\ref{secsec:upscaling}), and related research on language adaptation (§\ref{secsec:lang_adaptation}).

\subsection{Upscaling Techniques} 
\label{secsec:upscaling}

Model upscaling aims to reuse the learned behavior of small neural networks when training larger ones.

\citet{chen2016net2netacceleratinglearningknowledge} proposed Net2Net, a framework for function"-preserving widening and deepening of CNNs.
This idea was later adapted to Transformers by bert2BERT \citep{chen-etal-2022-bert2bert}, which reuses weights from the current and upper layer of the source model.
\citet{pmlr-v97-gong19a} applied a stacking approach to transfer knowledge from shallow to deeper BERT models, achieving similar accuracy with fewer training steps.
\citet{du2024stacking} showed that depth"-wise stacking (simply duplicating layers) offers the best speedup, while increasing width is less effective.
Tokenformer \citep{wang2025tokenformer} replaces a Transformer’s linear layers with attention between input tokens (queries) and parameter tokens (keys/values); scaling up then involves adding parameter tokens.

In this work, we use \emph{HyperCloning} \citep{pmlr-v262-samragh24a}, a symmetric method for layer expansion that scales up a weight matrix $W_S$ as
\begin{equation}
W_L \;=\; 
\begin{bmatrix}
\frac{W_S}{2}+\eta_1 & \frac{W_S}{2}-\eta_1 \\
\frac{W_S}{2}+\eta_2 & \frac{W_S}{2}-\eta_2 
\end{bmatrix}
\end{equation}
where $\eta_1$, $\eta_2$ are noise to break the symmetry in the continued training after upscaling.
Normalization layers and positional embeddings are expanded in the same way, while attention layers are scaled by increasing the number of heads.

\subsection{Language Adaptation}
\label{secsec:lang_adaptation}

We propose upscaling as a means of \emph{language adaptation} through continued pretraining of a base model. Concurrent work \citep{lee-etal-2026-scale} tests this idea on a single language, whereas we look at a diverse set of twelve target languages.

Continued pretraining is an established strategy for improving target"-language performance \citep{etxaniz-etal-2024-latxa, samuel-etal-2025-small}.
It is frequently implemented using parameter"-efficient finetuning techniques \citep{pfeiffer-etal-2020-mad, razumovskaia-lora, cui2024efficienteffectivetextencoding}.
While effective for larger models, full"-parameter finetuning can yield superior performance for smaller models \citep{yong-etal-2023-bloom}.

A challenge in continued pretraining is the catastrophic forgetting of previously learned languages \citep{gogolou2024continual}.
\citet{elhady-etal-2025-emergent} demonstrate that including English during continued pretraining is crucial not only for mitigating forgetting but also for preserving and enhancing capabilities in the target language.

\section{Experimental Setup}

We describe the training data (§\ref{secsec:data}), followed by our upscaling setup (§\ref{secsec:models}), our method for compute"-matched comparison of models (§\ref{secsec:flopmatched_setup}), and the multilingual models (§\ref{secsec:multilingual}). Finally, we summarize the evaluation datasets and procedures (§\ref{secsec:evaluation})\footnote{Models and datasets are available in this collection: \url{https://huggingface.co/collections/liu-nlp/grow-up-and-merge}. The code is available here: \url{https://github.com/liu-nlp/multilingual-scaling}.}.

\subsection{Data}
\label{secsec:data}

\paragraph{Languages}

We select a diverse set of languages representing different language families, scripts, morphological characteristics and resource levels. In total, we consider twelve target languages, presented in Table \ref{tab:training-data}.
The three Germanic languages (Swedish, Icelandic, and Faroese) are Indo-European, with Swedish typologically closest to English. Icelandic and Faroese are morphologically richer and, especially the latter, lesser"-resourced. 
Persian, Southern Pashto and Central Kurdish are also Indo-European but from the Indo-Iranian branch. Furthermore, they use Arabic-based scripts with minimal token overlap with the Latin-script languages, posing known challenges for multilingual models \citep{muller-etal-2021-unseen, liu-etal-2024-translico}.
The Uralic languages (Finnish, Estonian, and Northern Sámi) are agglutinative with case systems, with Northern Sámi being the least-resourced language in our experiments. The Niger-Congo languages (Shona, Swahili, and Zulu) are known for complex noun class systems. 

\paragraph{Corpora}

We use three different data sources for training our models: deduplicated FineWeb"-Edu \citep{lozhkov2024fineweb-edu} and Python"-Edu \citep{benallal2024smollmcorpus} for base model pretraining, and the training splits from FineWeb-2 \citep{penedo2025fineweb2pipelinescale} for continued pretraining on target languages.
We will refer to these as \CORPUS{English}, \CORPUS{Code} and \CORPUS{Multilingual data}. Table~\ref{tab:training-data} shows the number of documents for each data source, as well as the number of tokens when applying the Llama 3.3 tokenizer and the number of tokens used for training.

\begin{table}[t]
\centering\small
\begin{tabularx}{\linewidth}{lp{2.1cm}rrr}
\toprule
ISO3 & Dataset & \# Docs & \# Tokens & \# Seen \\
\midrule
& Code & 7.68M & 3.34B  \\
\texttt{eng} & English & 190M & 187B \\
\midrule
\texttt{ekk} & Estonian & 10.2M & 16.4B & 73B \\
\texttt{fin} & Finnish & 36.7M & 54.7B & 85.5B \\
\texttt{sme} & Northern Sámi & 65.7K & 74.9M & 449M \\
\texttt{fao} & Faroese & 291K & 230M & 1.4B  \\
\texttt{isl} & Icelandic & 3.01M & 4.3B & 25.8B \\
\texttt{swe} & Swedish & 59.5M & 64.2B & 64.2B \\
\texttt{ckb} & Central Kurdish & 555K & 1.04B & 6.2B \\
\texttt{fas} & Persian & 58.8M & 60.5B & 60.5B\\
\texttt{pbt} & Southern Pashto & 640K & 850M & 5.1B \\
\texttt{sna} & Shona & 84.4K & 136M & 816M \\
\texttt{swh} & Swahili & 1.21M & 1.24B & 7.4B \\
\texttt{zul} & Zulu & 127K & 200M & 1.2B \\
\bottomrule
\end{tabularx}
\caption{Overview of the training datasets. \emph{ISO3} refers to the three"-letter language code as per ISO-639-3. \emph{\# Tokens} refers to the size of the dataset whereas \emph{\# Seen} refers to the number of training tokens.}
\label{tab:training-data}
\end{table}

\paragraph{Setup}

We randomly split the English data into an 80\% and a 20\% subset, where the former is used to train a ``seed'' model, and the latter is set aside for experiments comparing continued pretraining and upscaling.
For target language adaptation, we combine the target"-language data with replay data, which has been shown to mitigate catastrophic forgetting \citep{scialom-etal-2022-fine}.
For replay, we use random subsets of the English and code data that were used for pretraining the seed model.
The amount of replay data is proportional to the amount of training data in the respective target language.
For Swedish, we use 1\% of the seed model's English data and 5\% of the code data; for other languages, we scale this down linearly based on the number of documents in that language (cf.\ Table\,\ref{tab:training-data}) compared to Swedish.
We also considered alternatives (e.g., leaving out code data), but found this setup to perform best in preliminary experiments. We exclude replay data for the three lowest"-resource languages (Northern Sámi, Shona and Zulu) with less than 0.5\% of the number of Swedish documents.

\subsection{Upscaling Experiments}
\label{secsec:models}

\paragraph{Architecture}

For all experiments, we adapt the SmolLM2 architecture \citep{allal2025smollm2smolgoesbig}.
Following their setup, our smallest models have 180M parameters due to using the heavily multilingual tokenizer of Llama~3.3 with a vocabulary size of 128K. For upscaling, we use HyperCloning \citep{pmlr-v262-samragh24a}, as presented in Section~\ref{secsec:upscaling}.
Since the input and output embeddings in our models are tied, we scale the output embedding matrix at runtime to match the output magnitude of the original model, following the reference implementation.\footnote{\url{https://github.com/apple/ml-hypercloning}} Our training framework is based on nanotron. \footnote{\url{https://github.com/huggingface/nanotron}}

\paragraph{Training Details}

We use a linear warm"-up of the learning rate during the first 0.2\% of the training steps (with a minimum of 10~steps where applicable) and a linear decay over the last 20\% of steps.
We train all models with a global batch size of 5,120 samples. For the \SETUP{1× cloned} models, we add noise and follow \citet{pmlr-v262-samragh24a} by applying a signal-to-noise ratio of 10 dB.
For all models in our experiments, the total combined training cost was 17,258 hours on NVIDIA A100 40GB GPUs. 

\paragraph{Base models}

Following the findings of \citet{aryabumi2024_tocodeornottocode}, who show that initializing language model training from a code"-pretrained checkpoint enhances reasoning capabilities, we first train a 180M"-parameter ``seed'' model on \CORPUS{Code} for two epochs, followed by training on a mix of \CORPUS{Code} and the 80\% \CORPUS{English} split for one epoch. From this initialisation, we derive two base models using the 20\% \CORPUS{English} split: (a)~a 180M"-parameter model, obtained by continued pretraining of the seed model as"-is, and (b)~a 572M"-parameter model, obtained by upscaling the seed model via HyperCloning with a scaling factor of 2 before continued pretraining.
This design ensures that both our base models have seen the same total amount of English data, allowing us to attribute any performance differences solely to scaling effects rather than training data variation. 

\paragraph{Target-language models}

We use three different setups for target language adaptation, illustrated in Figure~\ref{fig:scaling-setups}:
(1)~We continue pretraining the 180M"-parameter base model on the target"-language data (\SETUP{1×}).
(2)~We start from the same base model but scale it to 572M parameters before continuing pretraining on the target language (\SETUP{1× cloned}).
(3)~We continue pretraining the 572M"-parameter base model on the target"-language data (\SETUP{2×}).
Thus, the total amount of training data per language remains unchanged between the three setups. We ensure consistent data ordering between setups.

\subsection{Compute"-Matched Comparison}\label{secsec:flopmatched_setup}

In addition to final checkpoint scores, we analyze trajectories across checkpoints. To better understand training dynamics across model scales and scaling strategies, we calculate the total compute cost of each checkpoint, measured in hardware"-independent model FLOPs similar to previous work \citep{10.5555/3600270.3602446, kaplan2020scalinglawsneurallanguage}, including the pretraining cost of their corresponding base model. We record scores at intermediate checkpoints in 250"-step intervals, or approximately 2.6 billion tokens for most models. For models with fewer than 250 steps per epoch, we instead save a checkpoint at the end of each epoch. \citet{heineman2026signal} show that applying the exponential moving average (EMA) to training curves improves the decision accuracy of identifying the best of two models based on pairs of intermediate checkpoints during training. Following their result, we apply EMA over $N=20$ checkpoint windows to overcome high checkpoint-to-checkpoint noise, particularly in linguistic acceptability accuracies, and facilitate interpretation of training behavior.

\subsection{Multilingual Models}
\label{secsec:multilingual}

To investigate scaling dynamics during multilingual pre-training, we continuously pretrain three multilingual models. For that, we combine the data for each target language (including replay data) and follow the setups for \SETUP{1×}, \SETUP{1× cloned} and \SETUP{2×}.
We refer to these multilingual baselines as \SETUP{1× (multi)}, \SETUP{1× cloned (multi)}, and \SETUP{2× (multi)}.

To determine the total amount of training data for each language, we use a modified version of UniMax sampling \citep{chung2023unimax}, a strategy aiming for uniform coverage of larger languages while mitigating overfitting on smaller languages by setting a maximum number of epochs.
We set the UniMax character budget to 850.225 billion and the maximum number of epochs to~6, following a scaling law for data"-constrained models indicating that returns decrease quickly after more than 4~epochs \citep{muennighoff2023scaling}.
We continue pretraining our models for 1~epoch on Swedish and Persian, approximately 1.8~epochs on Finnish, 4.5~epochs on Estonian, and 6~epochs on all other languages.
Following the UniMax algorithm exactly with our character budget would have led to slightly more than 1~epoch for Swedish and Persian; to simplify comparisons, we fix the number of epochs for both languages to exactly~1 and re"-distribute the remaining character budget equally to Estonian and Finnish instead.

We include six multilingual pretrained models from previous work: Gemma~3 270M, Qwen~3.5 0.8B, Gemma~3 1B, Gemma~3 4B and Qwen~3.5 4B.
Both Gemma~3 \citep{gemmateam2025gemma3technicalreport} and Qwen~3.5 \citep{qwen3.5_blog} are highly multilingual, supporting more than 140 and 200 languages and dialects, respectively. The composition of languages is not public for the Gemma models, but Qwen 3.5 supports all of our target languages except for Northern Sámi.
All models included in our comparisons are base models without instruction tuning.

\subsection{Evaluation}
\label{secsec:evaluation}

\begin{figure*}
    \centering
    \includegraphics[width=1\textwidth]{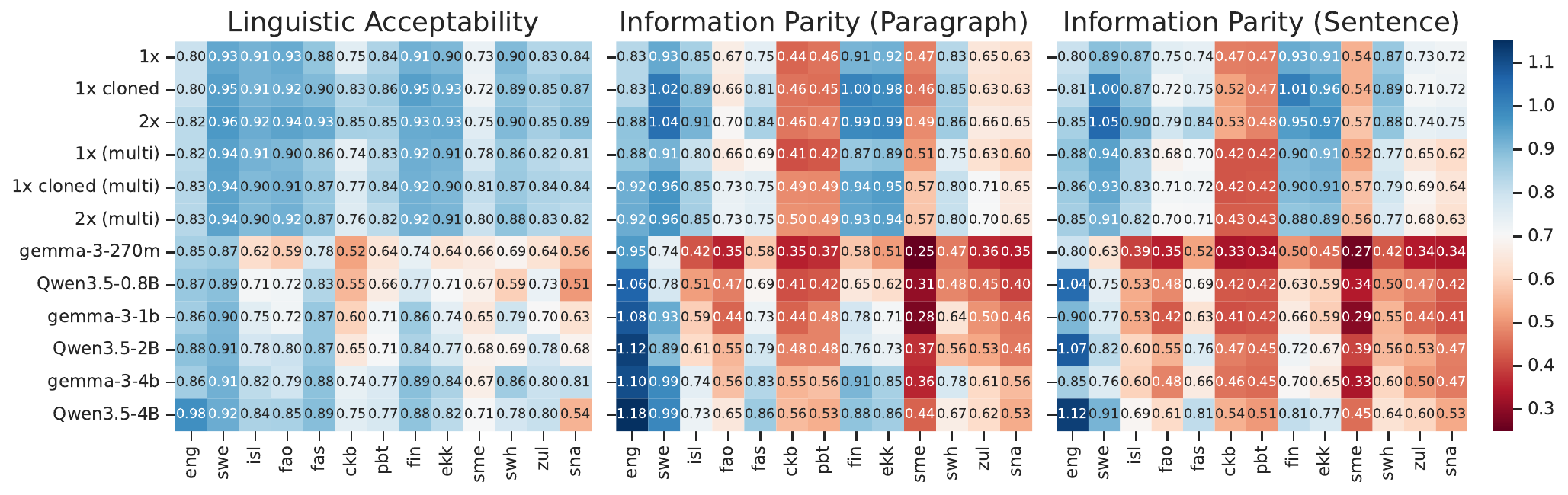}
    \caption{\label{fig:scaling}Results for all models and multilingual baselines at different scales. Linguistic acceptability scores are averages over the tasks introduced in Section~\ref{secsec:evaluation}. English scores of single"-target models are averaged across models.}
\end{figure*}

\paragraph{Information Parity (IP)}

Information Parity \citep{tsvetkov-kipnis-2024-information} is an intrinsic measure of multilingual capability that compares how efficiently a model compresses a target language relative to English, via a ratio of negative log"-likelihoods.
Values near parity (1) indicate similar encoding efficiency.
As IP correlates strongly with downstream performance \citep{tsvetkov-kipnis-2024-information}, it serves as a simple proxy for cross"-lingual generalization.
In the original definition of IP for multilingual models, the log"-likelihood of the English translation of a given text is taken from the same model as the target language log"-likelihoods.
However, for adaptation to a single target language, this would lead to IPs being artificially high when the English capabilities of the base model are partially lost.
Therefore, we use the log"-likelihoods of our largest English base model (\SETUP{2×}) as a reference to ensure consistent scores across languages, scales and training setups.
This allows us to also report IPs on English as an additional measure of catastrophic forgetting.
We compute IP using the BOUQuET dataset \citep{andrews-etal-2025-bouquet, omnilingual2026-bouquet}, which provides parallel sentence-aligned paragraphs across 275 languages.

\paragraph{Linguistic Acceptability (LA)}

Linguistic Acceptability probes quantify whether a model captures fine"-grained grammatical knowledge.
This is crucial for assessing whether adaptation yields robust morphosyntactic competence.
We evaluate LA using both existing manually annotated and custom automatically generated datasets.
All evaluations follow a minimal"-pair setup \citep{10.1162/tacl_a_00115}, in which each test item consists of one grammatically correct and one incorrect variant, and the model is expected to assign a higher probability to the correct form, and use accuracy over all pairs as the metric. 
We use two pre-existing benchmarks: English BLiMP \citep{warstadt-etal-2020-blimp-benchmark} and the multilingual MultiBLiMP \citep{jumelet2025multiblimp10massivelymultilingual}.
Inspired by MultiBLiMP, we construct a more diverse (in phenomena covered) set of language"-specific BLiMP"-style datasets for morphology through systematic perturbations based on UniMorph annotations.
We sample sentences from high"-quality Wikipedia articles (e.g., articles tagged as \emph{excellent} or \emph{article of the month}) and create minimal pairs by replacing a single word with an alternative form that differs in exactly one UniMorph feature (e.g., person, gender, or case).
Table~\ref{tab:language-subsets} in Appendix~\ref{appendix: blimp} provides detailed statistics for each dataset.

\paragraph{Statistical significance testing}

We use paired bootstrap resampling to evaluate the statistical significance of the observed differences between the three different setups for target-language adaptation. For each of the three pairwise model comparisons, we collect 10,000 resamples per language and calculate the two-sided $p$-value as
\begin{displaymath}
    p = 2 \min\left(P(\Delta \leq 0), P(\Delta \geq 0)\right)\,.
\end{displaymath}
Finally, we apply the Benjamini"=Hochberg correction \citep{benjamini_hochberg} within each pairwise comparison, across languages. We report detailed results in Appendix \ref{appendix:significance_tests}.

\begin{figure*}
    \centering
    \includegraphics[width=1\textwidth]{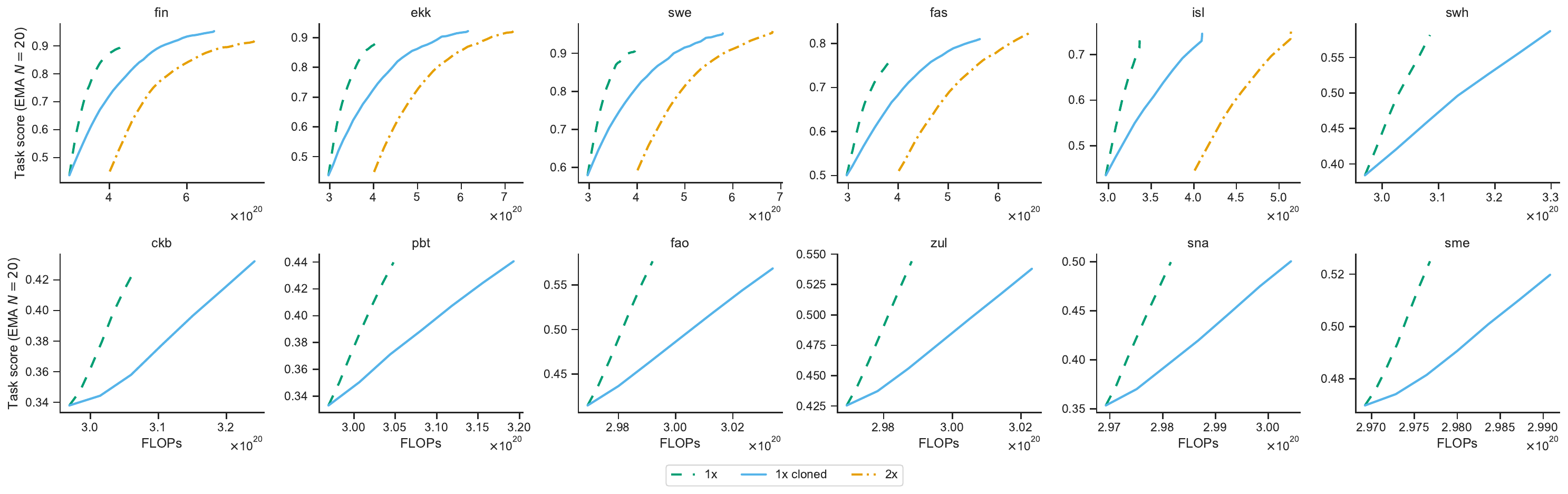}
    \caption{\label{fig:monolingual}Training curves for models continually pre-trained on a single target language by total training FLOPs, incl.\ base models. Scores are averaged over all sub-tasks and smoothed with EMA over a window of 20 checkpoints.}
\end{figure*}

\begin{figure}
    \centering
    \includegraphics[width=1\linewidth]{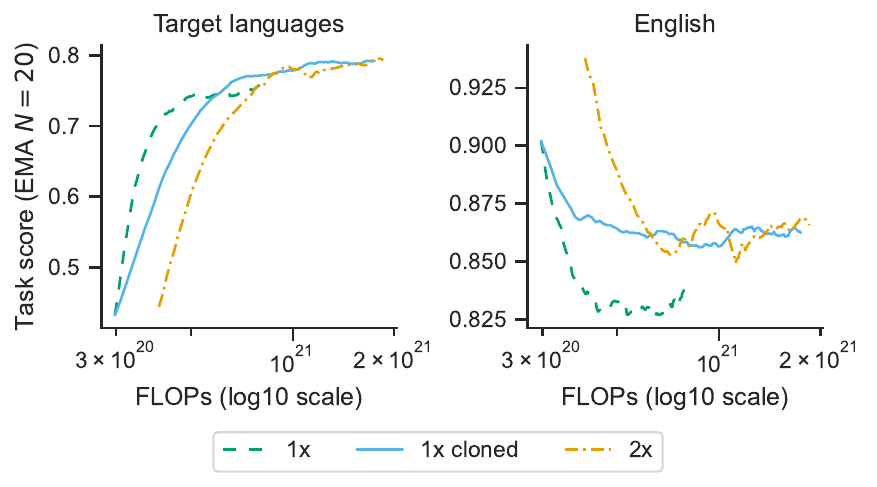}
    \caption{\label{fig:multilingual}Training curves for models continually pre-trained on our multilingual data mixture by total training FLOPs, incl.\ base models. Scores are averaged over sub-tasks and smoothed (EMA; window of 20 checkpoints).}
\end{figure}

\section{Results}
\label{sec:results}

We start by comparing our single"-target and multilingual models across scaling setups on the same amount of training data (§\ref{secsec:results_data}).
Thereafter, we analyze training dynamics in the compute"-matched setting, first for single-target models (§\ref{secsec:results_monolingual}) and then for multilingual models (§\ref{secsec:results_multilingual}). 
Finally, we probe the internal representations of the multilingual models for language and meaning discrimination (§\ref{sec:abx}). Furthermore, to complement and further support the intrinsic evaluation measures reported in this section, we report downstream evaluation results on the SIB-200 topic classification dataset \citep{adelani2023sib200} in Appendix~\ref{app:sib200}.

\subsection{Data-Matched Comparisons}
\label{secsec:results_data}

Figure~\ref{fig:scaling} presents the results for training our single"-target and multilingual models across different strategies on the same amount of data, alongside strong multilingual baselines from prior work.
Our pairwise significance tests (Appendix~\ref{appendix:significance_tests}) show that all absolute IP differences greater than $0{.}005$ between our data-matched models are significant (adjusted $p < 0.05$) under bootstrap resampling.

\paragraph{Data efficiency}

All of our monolingual and multilingual models achieve comparable or higher scores on LA and IP compared to the baselines on most target languages, including the largest among them, Qwen3.5 4B and Gemma 3 4B, with over 22 times more parameters than our \SETUP{1×} models. The exceptions are the IP of the three Indo-Iranian languages and Swedish, as well as Persian LA, for which our \SETUP{1×} models score higher than baselines at or below one billion parameters. On English, all baselines outperform our continuously pre-trained models.
LA scores show only small differences across model scales for our models, but are substantially higher than the scores of the massively multilingual baseline models. This suggests that target"-language adaptation is beneficial compared to using a general multilingual model, but that the model capacity necessary to capture syntactic and morphological preferences is relatively small for a specialized model.
For the longer contexts in paragraph"-level IP, however, both the \SETUP{2×} models and the multilingual and high"-resource single"-target \SETUP{1× cloned} upscaled models clearly outperform the \SETUP{1×} models, showing that the larger models can handle long contexts more effectively.

\paragraph{Multilingual and single"-target} Compared to single"-target models of the same scale, our \emph{multilingual} models generally underperform on LA and sentence-level IP across most languages. However, on paragraph-level IP the multilingual \SETUP{2×} and \SETUP{1× cloned} models often outperform for the lower-resource languages in our study, indicating some positive multilingual transfer in this higher"-context setting. Multilingual training is most consistently beneficial for our lowest"-resource language, Northern Sámi, and for English across all metrics.

Comparing our scaling strategies in the single"-target setting, \SETUP{2×} is overall more data"-efficient than \SETUP{1× cloned}, particularly in IP and for higher"-resourced languages, with the exception of Finnish. \SETUP{1× cloned} models are still more data"-efficient or comparable to \SETUP{1×} models for higher"-resourced languages, but reach lower scores for some of our lowest"-resource languages. Overall, \SETUP{2×} most consistently outperforms \SETUP{1×} across the languages we consider.

Our multilingual \SETUP{1× cloned} and \SETUP{2×} models perform mostly identically on average, with differences of no more than $0.03$ for the individual languages across all metrics, and less than $0.01$ for paragraph-level IP. The multilingual \SETUP{1×} model scores close to identically on sentence-level IP and LA tasks. However, we see a clear benefit of scale in the longer context paragraph-level IP scores, in which multilingual \SETUP{2×} and \SETUP{1× cloned} substantially outperform \SETUP{1×} across languages.

Average English scores are nearly identical between \SETUP{1×} and \SETUP{1× cloned} but higher in \SETUP{2×} single"-target models. In contrast, \SETUP{1× cloned} and \SETUP{2×} English scores are almost identical and higher than \SETUP{1×} scores in multilingual models. The exceptions are sentence-level IP scores, which are higher for \SETUP{1×} models. We further analyze the dynamics of English in multilingual models in §\ref{secsec:results_multilingual}.

\subsection{Compute-Matched \qquad\qquad\qquad\qquad Single Target Language Scaling Dynamics} \label{secsec:results_monolingual}

When analyzing evaluation scores during training, matched for FLOPs across single-target models (Figure~\ref{fig:monolingual}), we find that \SETUP{1×} setups are generally more compute"-efficient than \SETUP{1× cloned} as they can be trained on more data at substantially lower cost. However, given a sufficient amount of training data, \SETUP{1× cloned} becomes more data-efficient.
While \SETUP{2×} models initially converge faster when matched for training data, \SETUP{1× cloned} models follow similar learning trajectories. Combined with the lower initial cost of their base models, this results in increased compute efficiency for higher-resource languages. However, as discussed in §\ref{secsec:results_data}, \SETUP{2×} is more data-efficient for low-resource languages and some higher-resource languages such as Persian.

\subsection{Compute-Matched \qquad\qquad\qquad Multilingual Scaling Dynamics}
\label{secsec:results_multilingual}

We compare scaling strategies for training multilingual models relative to total training FLOPs separately on an average of target language scores and English scores (Figure~\ref{fig:multilingual}).
As in our single"-target experiments, \SETUP{1×} initially reaches higher scores at a lower compute budget but is overtaken by \SETUP{1× cloned} approximately halfway through training, and then by \SETUP{2×} by the end of training. As established in Section~\ref{secsec:results_data}, \SETUP{1× cloned} and \SETUP{2×} show comparable levels of data efficiency. By skipping the initial English upscaling step, \SETUP{1× cloned} reaches these comparable average scores $\approx 1.04\times10^{20}$ FLOPs earlier.
On English, we find that all models show signs of catastrophic forgetting. \SETUP{1×} reaches the lowest scores on average. The average score of \SETUP{1× cloned} decreases more slowly. The \SETUP{2×} model starts at a higher average score due to the increased capacity of its English base model and initially loses English capabilities more slowly, but reaches a similar level as \SETUP{1× cloned} by the end of training.

\subsection{Language and Meaning Discrimination}
\label{sec:abx}

\begin{figure}
\centering
    \includegraphics[width=\columnwidth]{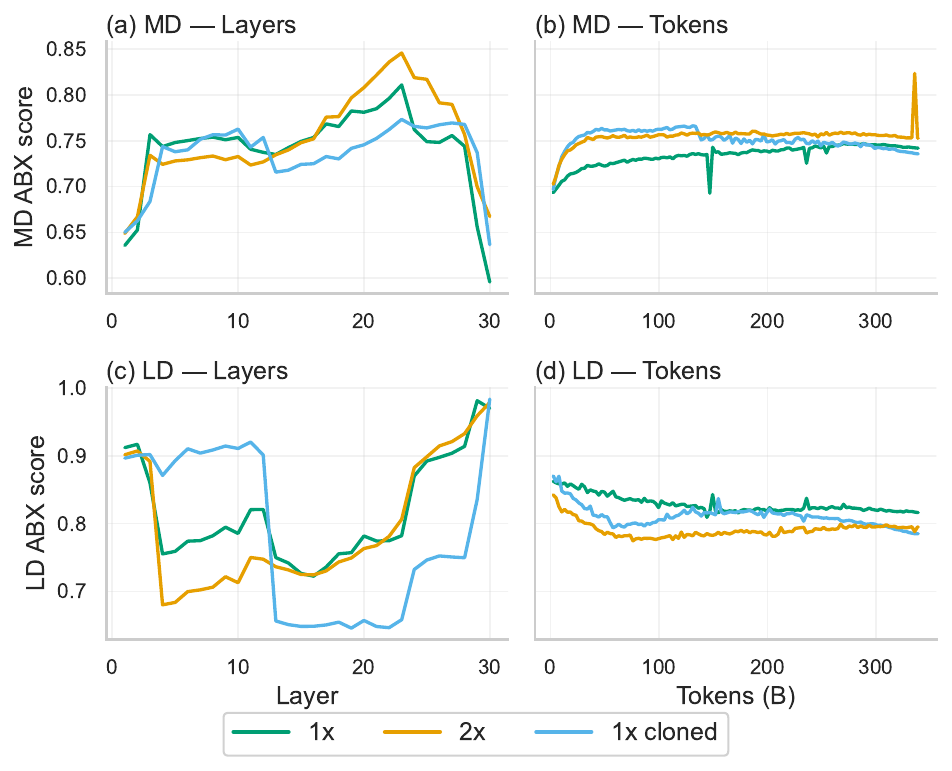}
    \caption{ABX scores for our multilingual models. All scores are averaged over language pairs; checkpoint scores are also averaged over layers. }
    \label{fig:abx_layer_cp}
\end{figure}

To gain further insight into the differences in training dynamics between standard continued pre-training and hypercloning initializations, as well as into the representations across layers, we analyze the multilingual alignment of model representations across intermediate checkpoints and layers. 
We apply the ABX task proposed by \citet{de-seyssel-etal-2025-discriminating} using parallel sentences from the BOUQuET dataset \citep{andrews-etal-2025-bouquet, omnilingual2026-bouquet} to investigate how language and meaning differentiation evolve during training in our models. 
ABX probes use triplets of sentences $(X, A, B)$ and compute the proportion of triplets for which the representation of $X$ is closer to $A$ than to $B$ under cosine distance.
Like \citet{de-seyssel-etal-2025-discriminating}, we employ two tasks:
In \emph{Meaning Discrimination (MD)}, for each language pair $(\ell_1, \ell_2)$, triplets are constructed such that $X$ is a sentence in~$\ell_1$, $A$ is the translation of~$X$ into~$\ell_2$, and $B$~is a different sentence in $\ell_2$. A high MD score indicates that representations of translation equivalents are more similar to each other than to unrelated sentences in the same language, i.e.\ that meaning is encoded cross-lingually. For \emph{Language Discrimination (LD)}, triplets are constructed such that $X$ is a sentence in~$\ell_1$, $A$ is a different sentence in the same language, and $B$~is the translation of~$A$ into~$\ell_2$. A high LD score indicates that same-language representations cluster together more.

We use ABX probes to study two questions: (1) how the representations change during training, and (2) how the representations are organized across layers in the final model. We compare the three multilingual models (\SETUP{1×}, \SETUP{1× cloned}, and \SETUP{2×}) across both checkpoints and layers\footnote{Appendix~\ref{app:probing_results} contains additional results on monolingual models and ablations across languages and groupings. }. 
Figure~\ref{fig:abx_layer_cp} reveals that the three models show clear differences in how their representations are learned and organized. 

\paragraph{Across layers}

\textbf{MD} scores increase quickly in the lower layers, peak in the middle layers, and decrease again toward the final layers.
This matches earlier work showing that middle layers encode more language"-neutral semantic information \citep{tenney-etal-2019-bert, jawahar-etal-2019-bert}.
The \SETUP{2×} and \SETUP{1× cloned} models behave similarly in the early layers, but begin to differ around layers 12--13.
The \SETUP{2×} model continues improving and reaches the highest MD scores overall.
In contrast, the \SETUP{1× cloned} model drops strongly in the middle layers and recovers only partly in later layers.
The \SETUP{1×} model has lower MD scores throughout.
\textbf{LD} scores show the opposite pattern.
They are highest in the first and last layers, where more token"-level and language"-specific information is processed.
The \SETUP{1×} and \SETUP{2×} models reduce LD quickly in the early layers.
The \SETUP{1× cloned} model behaves differently: it keeps higher LD for longer, then drops sharply around layers 12--13 (the same region where its MD scores decline) and stays below the other models until the final layers.
So, although \SETUP{2×} and \SETUP{1× cloned} reach very similar performance in evaluations, it appears that the representational space of the \SETUP{2×} model is more semantically aligned across languages.
In contrast to the other models, the \SETUP{1× cloned} model has a \emph{more} language"-distinct representation in the first 12 layers, but a \emph{less} language"-distinct representation in layers 13--28.
An ablation in Appendix~\ref{app:probing_results} (Figure~\ref{fig:ld_ablation}) shows that this happens independently of script or language family differences.

\paragraph{Across checkpoints}

The checkpoint analysis shows how these differences develop during training.
For \textbf{MD}, early in training, \SETUP{1× cloned} model scores highest, followed by \SETUP{2×} and then \SETUP{1×}. Around 135--145B tokens, however, MD in the \SETUP{1× cloned} model starts to \emph{decrease} (as the only model) and eventually falls below both other models.
A closer layer"-wise analysis (Appendix~\ref{app:probing_results}, Figure~\ref{fig:abx_layerwise_dynamics}) shows that the decline mainly happens in the middle layers, while the final layer is less affected, which explains why the models do not differ much in their final \emph{performance}, but only in their \emph{representational space}.
The \textbf{LD} scores show that all models become less language"-discriminative during training, but with different dynamics.
The \SETUP{1×} model decreases steadily,
the \SETUP{2×} model first decreases and then partially recovers, and 
the \SETUP{1× cloned} model fluctuates more strongly.
By the end of training, the \SETUP{1× cloned} and \SETUP{2×} models reach similar LD levels, while the \SETUP{1×} model retains the most language"-specific representations.
Overall, we observe that \SETUP{1× cloned} shows the most unstable behavior, which can also be traced to the middle layers (Appendix~\ref{app:probing_results}).

\section{Discussion}

The results from Section~\ref{sec:results} demonstrate that scaling via HyperCloning improves target"-language performance over regular continuous pre-training given the same data, supported by the statistical significance tests presented in  Appendix~\ref{appendix:significance_tests}.
Both our single"-target and multilingual models outperform heavily multilingual state"-of"-the"-art baselines with as much as 22 times more parameters on eight out of thirteen languages, showing that heavy multilinguality still incurs a cost, particularly for lower"-resource languages.

When comparing \SETUP{1× cloned} and \SETUP{2×} strategies for continued pre-training for single target languages, we find that \SETUP{1× cloned} is generally more compute-efficient when factoring in the additional cost of upscaling on English first. However, data efficiency varies between languages. \SETUP{2×} is generally equivalent or more data-efficient than \SETUP{1× cloned} except for Finnish. This is particularly the case for some of the lowest resource languages in our comparisons, such as Northern Sámi, Shona, and Zulu, where \SETUP{1× cloned} models are not trained long enough to optimally utilize their newly increased parameter space. For limited compute budgets, \SETUP{1×} models are generally more efficient as they can ingest more data with fewer FLOPs at the cost of lower data efficiency. This suggests \SETUP{1×} could still be considered in highly compute"-constrained environments, particularly when data is plentiful.

Moreover, particularly when adapting larger models, scaling may not be desirable due to the increase in inference cost when deployed. In contrast, as established in language model scaling law literature \citep{10.5555/3600270.3602446, kaplan2020scalinglawsneurallanguage}, the smaller \SETUP{1×} models hit a point of diminishing returns earlier, particularly in high-resource settings. Therefore, where data efficiency is critical, such as for low-resource languages, we show that investing additional compute into scaling in a high resource base model's language first can be more data efficient. However, the degree of improvement varies by target language.

\pagebreak

When training multilingual models, we find that upscaling directly on the multilingual data outperforms continuously pre-training from our smaller model and performs almost identically to upscaling on English first, making it the ideal scaling setup for increasing multilingual capacity in our experiments. Furthermore, while English forgetting occurs in all models in our setup, \SETUP{1× cloned} models show less forgetting than \SETUP{1×} models and similar final English scores to \SETUP{2×} at the end of training, despite the scaled base model of \SETUP{2×} being more proficient in English initially. The greatest differences between model scales were revealed by paragraph"-level information parity scores, showing the utility of longer documents in BOUQuET and similar translation datasets for more robust and discriminative evaluation of multilingual models.

Through an analysis of the internal representations of the multilingual models, we find evidence of higher alignment and less language"-dependent activations across languages in the \SETUP{2×} model. While initially following a similar trend during training, alignment scores in \SETUP{1× cloned} start fluctuating and ultimately drop to the lower level of \SETUP{1×}, indicating a lower representational stability of \SETUP{1× cloned} that is not reflected in task scores. Further work is needed to investigate potential impacts on, e.g., cross"-lingual generalization.

\section{Conclusion}

In this paper, we evaluated upscaling as a strategy for training and adapting models to new target languages. We found that upscaling the base model overcomes capacity constraints and improves target-language performance, particularly for high-resource languages and when training multilingual models. The choice of upscaling strategy depends on the use case: upscaling via English first is primarily advantageous for low-resource languages. Upscaling directly on target-language data is more compute-efficient when sufficient target-language data is available at the cost of data efficiency for some languages.

Furthermore, we show that upscaling on data from multiple target languages directly, rather than scaling on English first, is the ideal setup for multilingual models both in terms of data efficiency and reduced English forgetting. Through an analysis of the representation space, we find higher multilingual alignment when scaling on English first but lower stability when directly scaling on target languages, warranting further analysis of the mechanisms and impacts of this effect in future work. While we present a carefully controlled study of adaptation from a monolingual model, future work could investigate scaling as a means to add languages to or reduce language interference in pre-trained multilingual models.

\section*{Limitations}

We only study the upscaling behavior of models on twelve target languages due to compute constraints. For the same reason, we perform experiments only for a limited number of model sizes, and we can thus only draw conclusions regarding these small sizes. Thus, future work is needed to study the effects of target"-language upscaling for a broader range of languages and for larger models. Lastly, we use pretraining datasets published by other researchers due to time and budget constraints. While it was possible to choose a corpus of English that presumably does not contain unethical material due to its educational content, the choice of available large-scale corpora for our target languages is more limited and might contain inappropriate content.

\section*{Acknowledgments}

This work was supported by the Wallenberg AI, Autonomous Systems and Software Program (WASP) funded by the Knut and Alice Wallenberg Foundation, by TrustLLM funded by Horizon Europe GA 101135671, and by the Graduate School in Computer Science (CUGS). The computations were enabled by the Berzelius resource provided by the Knut and Alice Wallenberg Foundation at the National Supercomputer Centre and by the National Academic Infrastructure for Supercomputing in Sweden (NAISS), partially funded by the Swedish Research Council through grant agreement no. 2022-06725.

\bibliography{custom}

\appendix

\setcounter{bottomnumber}{3}

\section{UniMorph BLiMP Dataset} \label{appendix: blimp}

We provide details on our custom UniMorph BLiMP dataset in Table~\ref{tab:language-subsets}: Number of samples, number of subsets, and comments on where we did manual improvements on known weaknesses of either the UniMorph in the given language or of our automatic approach of creating the datasets. We release the dataset under the  Creative Commons Attribution-ShareAlike 4.0 International License (CC BY-SA), which we inherit from Wikipedia. 

\begin{table*}[h]
\centering
\begin{tabularx}{\textwidth}{lcc X}
\toprule
Language & Rows & Subsets & Comments \\
\midrule
ckb &   907 &  11 & -- \\
ekk &  5182 &  28 & Uses \textsc{Eesthetic} \citep{beniamine2024eesthetic} instead of UniMorph for broader and higher-quality coverage. \\
fao &  7907 &  40 & -- \\
fas &  1584 &   8 & UniMorph entries post-edited; missing non-compound verbs added. \\
fin & 10033 &  64 & -- \\
isl &  5200 &  26 & -- \\
pus &  3211 &  19 & -- \\
sna & 397 & 10 & -- \\
sme & 6,125 & 47 & -- \\
swe &  2000 &  10 & Archaic forms filtered out. \\
swh & 527 & 8 & -- \\
zul & 937 & 11 & -- \\
\bottomrule
\end{tabularx}
\caption{Overview of the custom BLiMP datasets. \# Subsets refers to the number of categories of morphological corruptions included in the dataset.
}
\label{tab:language-subsets}
\end{table*}

\section{SIB-200 Fine-Tuning}
\label{app:sib200}

\begin{figure*}
    \centering
    \includegraphics[width=\textwidth]{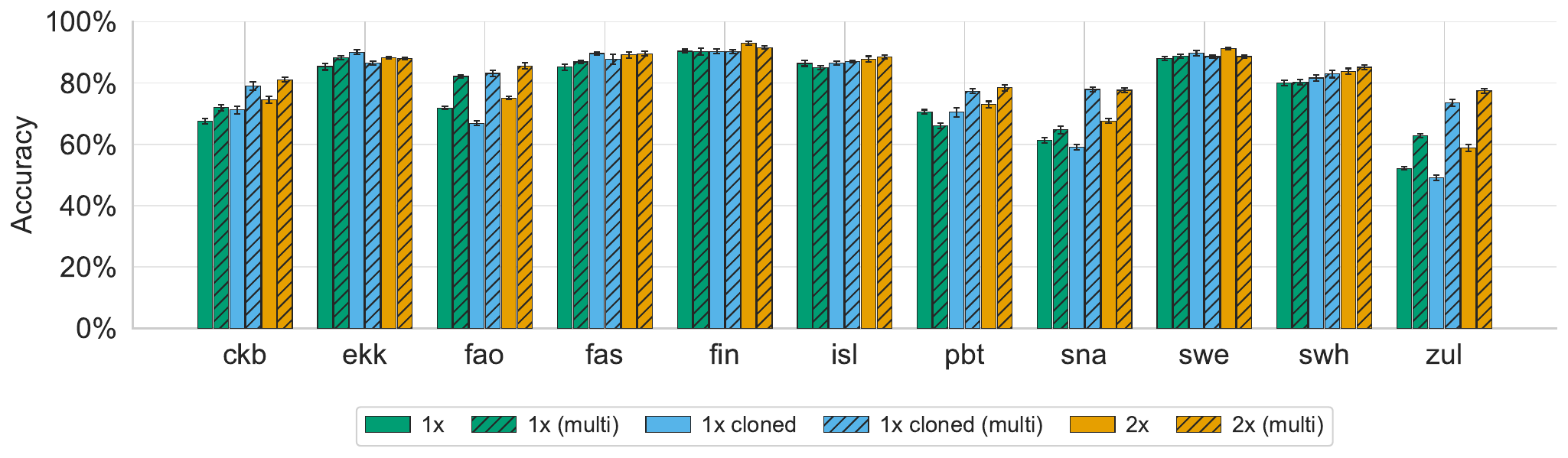}
    \caption{SIB-200 fine-tuning using a classification head.}
    \label{fig:sib200_by_language}
\end{figure*}

To test the models' performance under task-specific fine-tuning, we fine-tune all models on the topic classification dataset SIB-200 \citep{adelani2023sib200}, using the original train–development–test split. To control for differences in the number of trainable parameters during task-specific fine-tuning, we always add a classification head with a hidden size of 256 on top of the frozen model models. 
We use a learning rate of 5e-4, selection of the best checkpoint over 20 epochs, an effective batch size of 16, a cosine learning rate schedule with a warmup ratio of 0.05.  We train all models with 10 seeds and report mean and standard deviation. 
Results are in Figure~\ref{fig:sib200_by_language}. 
\paragraph{Scales} We see that the \SETUP{2×} models consistently outperform the \SETUP{1×} models, confirming that upscaling benefits target language performance. 
The role of the \SETUP{1× cloned} models is less clear: For the monolingual models and the lower-resource languages, they often underperform \SETUP{1×}. This is in line with our findings from Section~\ref{secsec:results_monolingual} where \SETUP{1× cloned} also underperformed for smaller languages. 
For the larger languages however, the \SETUP{1× cloned} models are sometimes even better than \SETUP{2×}. 

\paragraph{Multilingual versus Bilingual Models}
For the lower-resource languages (Central Kurdish, Southern Pashto, Faroese, Shona and Zulu), the multilingual models outperform their single-target counterparts by a wide margin. This contrasts with the intrinsic evaluations in Section~\ref{secsec:results_multilingual},
where multilingual models scored similarly or below single-target models of the same scale. We conclude that the multilingual models, trained on more data, are a better base for fine-tuning, although they underperform out of the box as the single-target models are more specialized. Topic classification relies less on fine-grained morphosyntactic knowledge than linguistic acceptability and more on robust semantic representations, which the lowest-resource languages appear to acquire through transfer from other languages in the mixture. 

\section{Extended ABX Results}
\label{app:probing_results}

In the main paper, we focused on the multilingual models under the different scaling setups. Here, we provide additional analyses of the ABX results, including layer-wise behavior, training dynamics, and per-language breakdowns.

\begin{figure}
    \centering
    \includegraphics[width=0.99\linewidth]{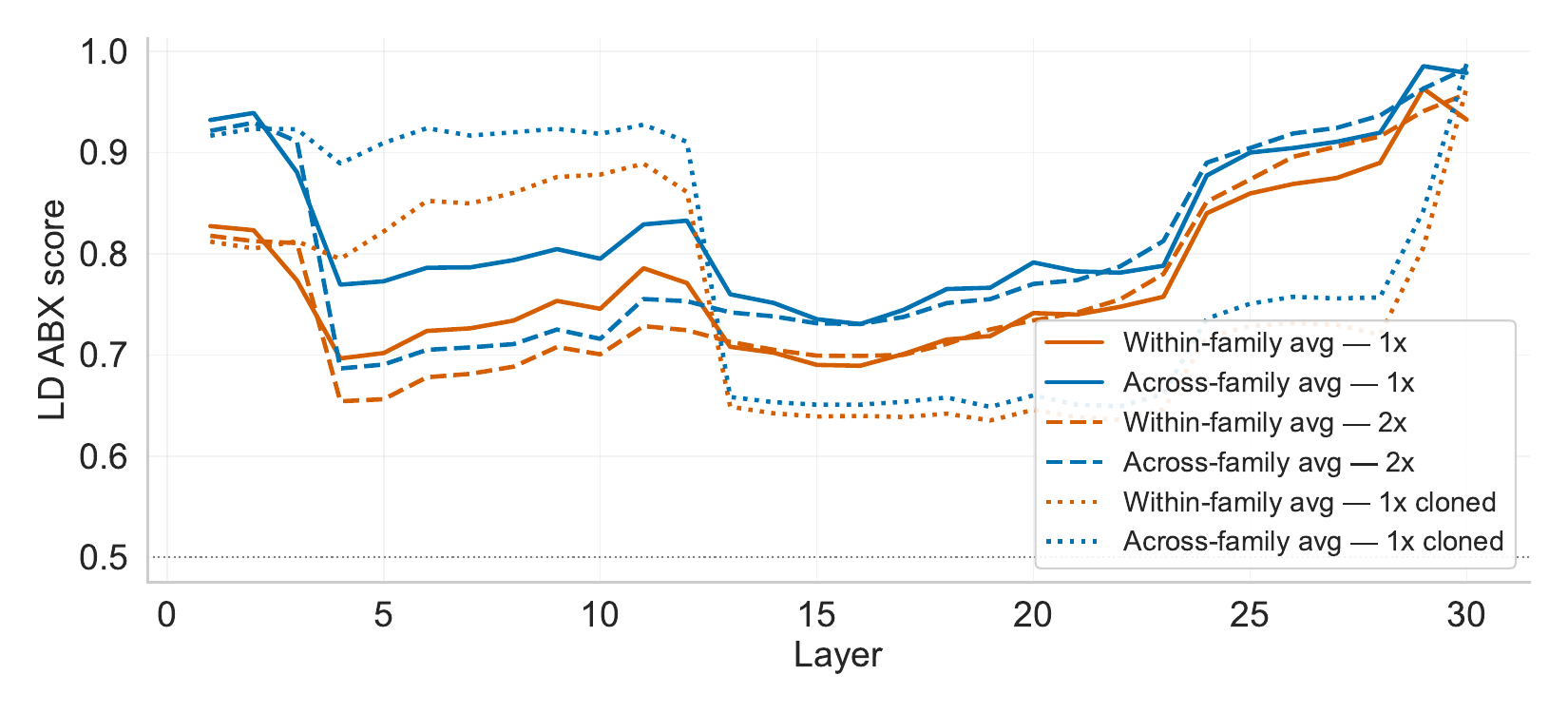}
    \includegraphics[width=0.99\linewidth]{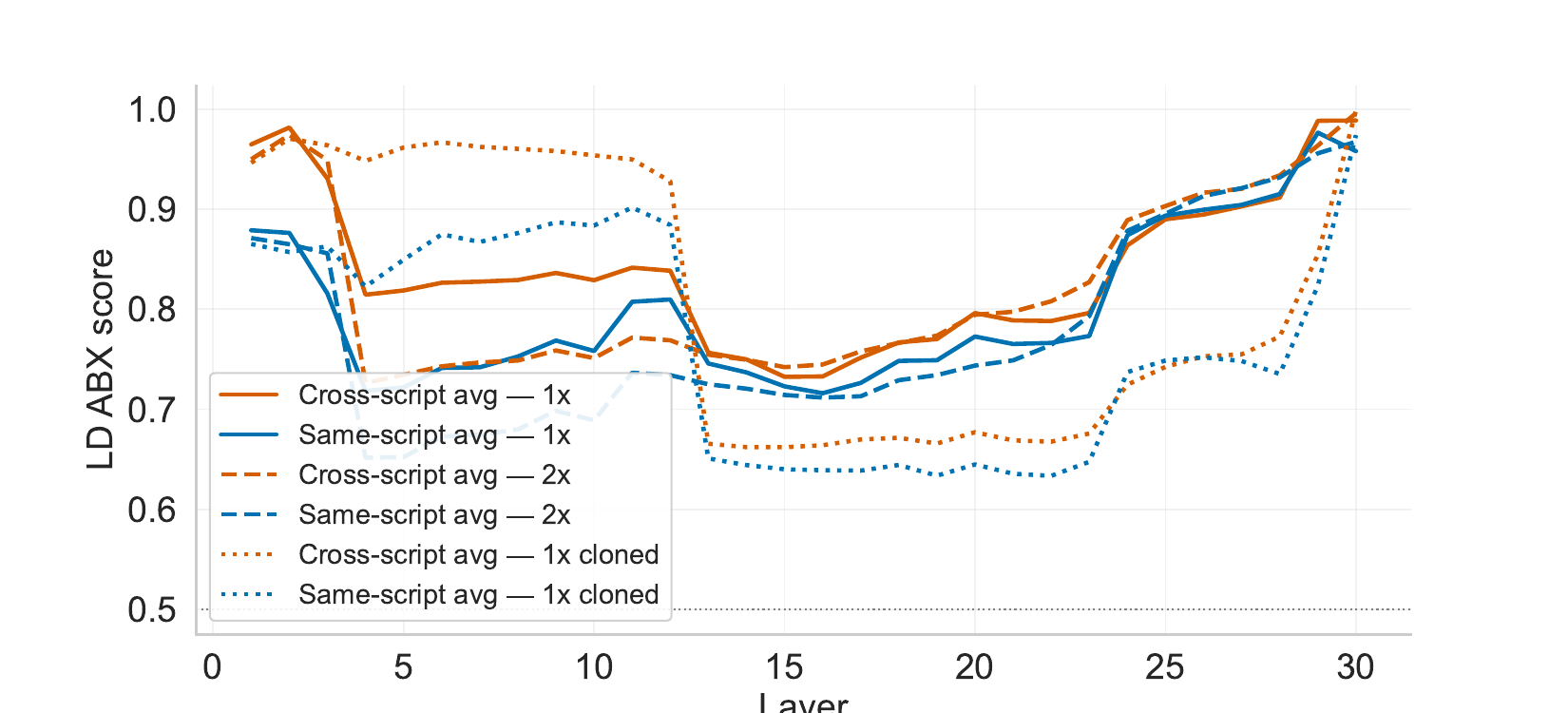}
    \includegraphics[width=0.99\linewidth]{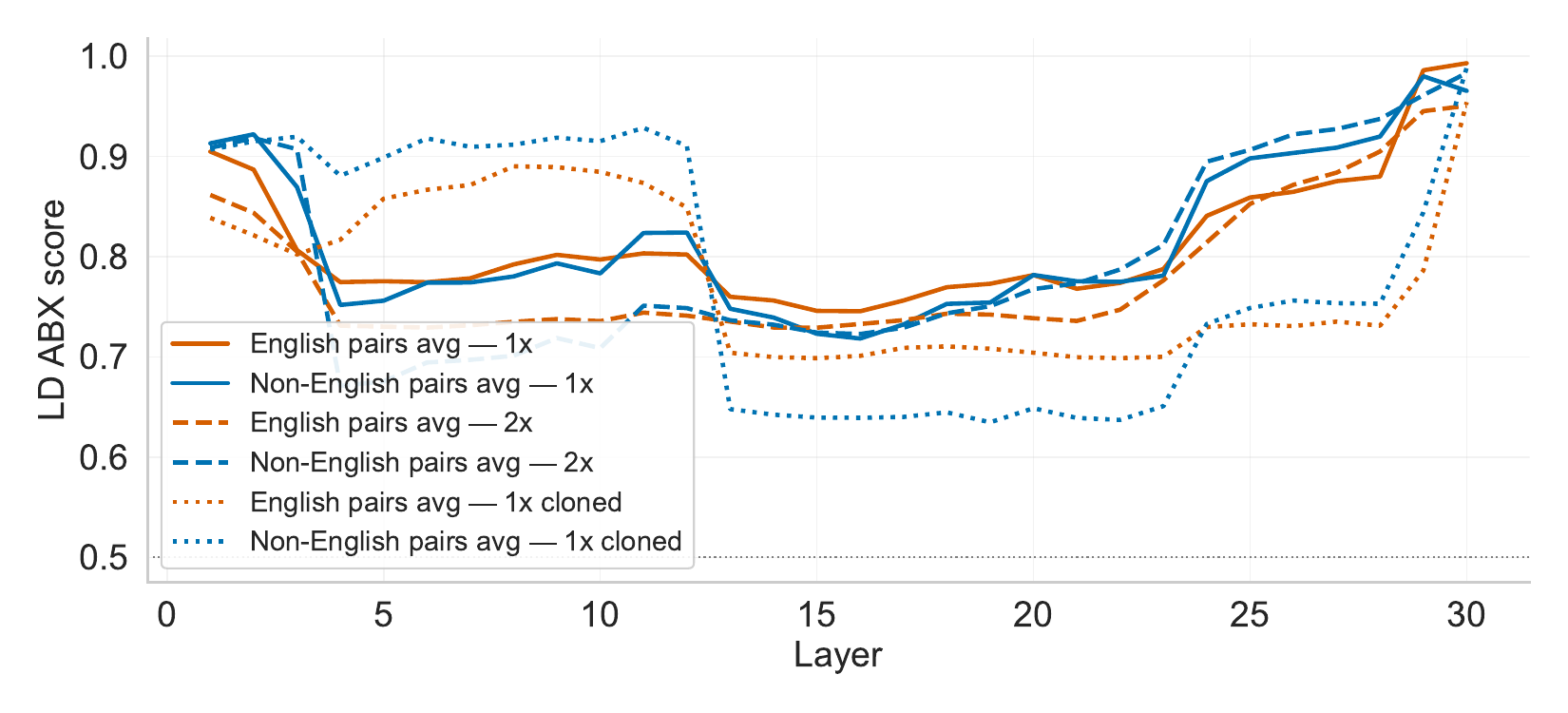}
    \caption{Ablation: LD ABX scores across layers for the three multilingual models, broken down by language pair type (within-family vs.\ cross-family, same-script vs.\ different-script, English vs.\ non-English pairs).}
    \label{fig:ld_ablation}
\end{figure}

\paragraph{Per-layer analysis} We recall from Figure~\ref{fig:abx_layer_cp} that MD scores rise rapidly in the lower layers, plateau in the middle layers, and decrease again in the uppermost layers. While \SETUP{2×} and \SETUP{1× cloned} initially behave similarly, the \SETUP{1× cloned} model begins to decline around layer 12 and recovers only slowly, whereas \SETUP{2×} continues improving. The \SETUP{1×} model consistently obtains lower MD scores.
LD exhibits the inverse trend. Scores are highest in the first and last layers, indicating stronger language-specific information. The \SETUP{1×} and \SETUP{2×} models show a rapid early decrease in LD, whereas \SETUP{1× cloned} preserves higher LD for longer before dropping sharply around layers 12--13. It then remains below the other models until the final layers.
To better understand this behavior, Figure~\ref{fig:ld_ablation} breaks the LD results down by language pair type: within-family versus cross-family, same-script versus different-script, and English versus non-English pairs. The same overall pattern appears in all settings, suggesting that the unusual behavior of the \SETUP{1× cloned} model is not driven by one specific group of language pairs. Some groups exhibit the pattern stronger than others (in particular, pairs not including English more strongly than pairs including English, and across-family pairs more strongly than within-family pairs), but it is present in all groups analyzed, suggesting a general effect. 

\begin{figure}
    \centering
    \includegraphics[width=0.99\linewidth]{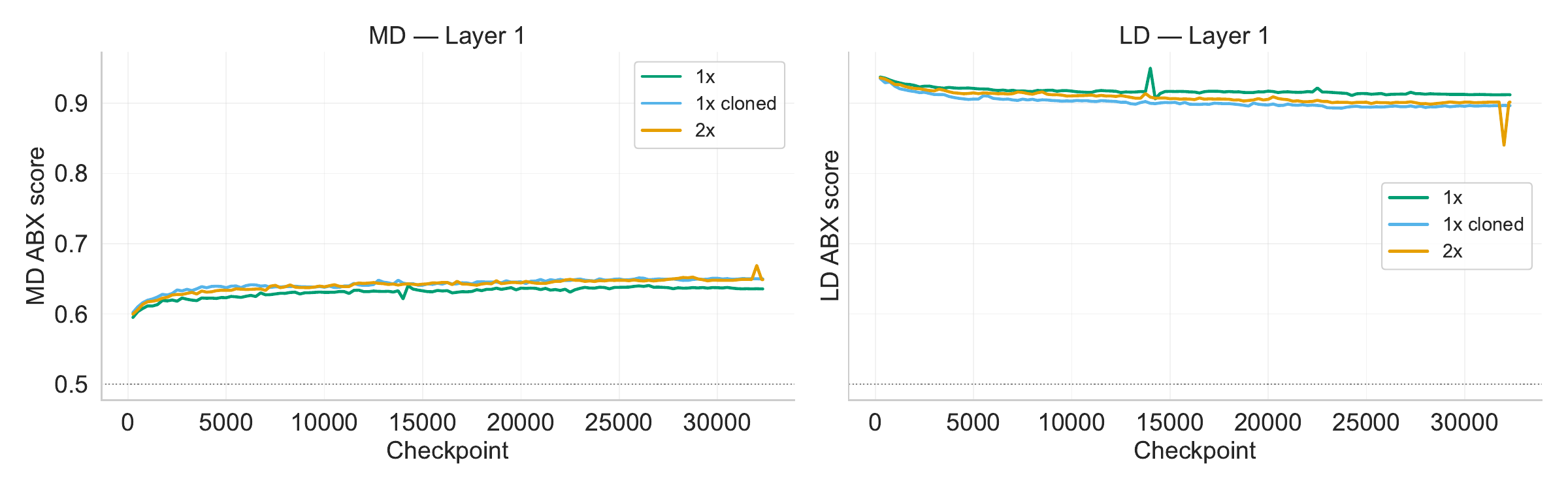}
    \includegraphics[width=0.99\linewidth]{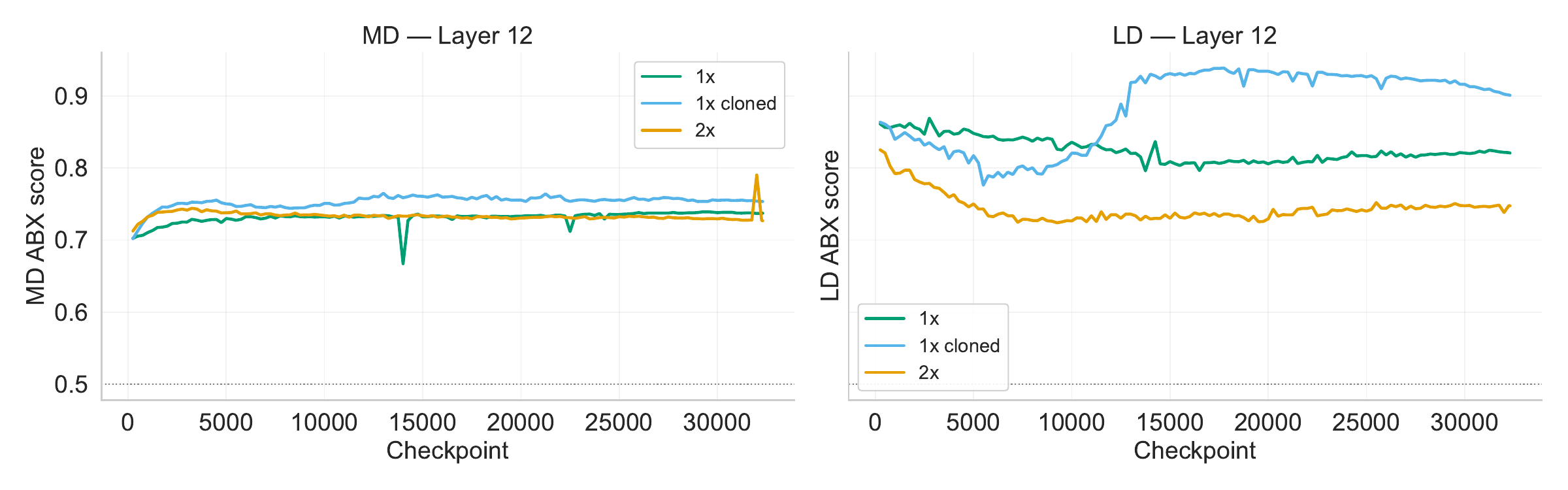}
    \includegraphics[width=0.99\linewidth]{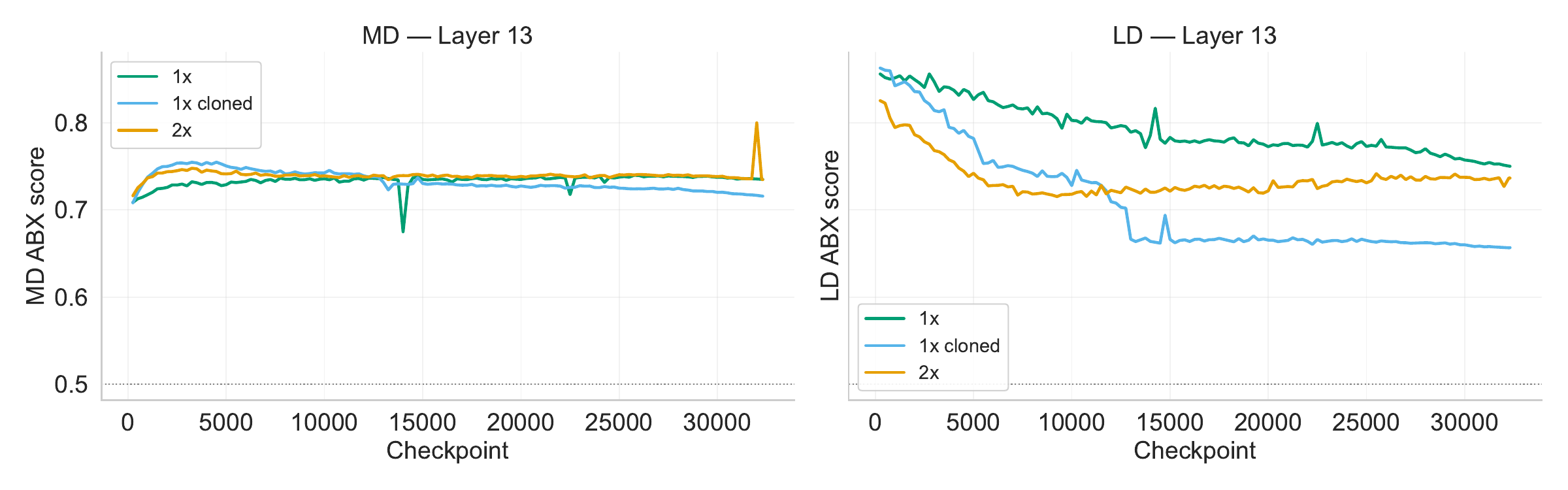}
    \includegraphics[width=0.99\linewidth]{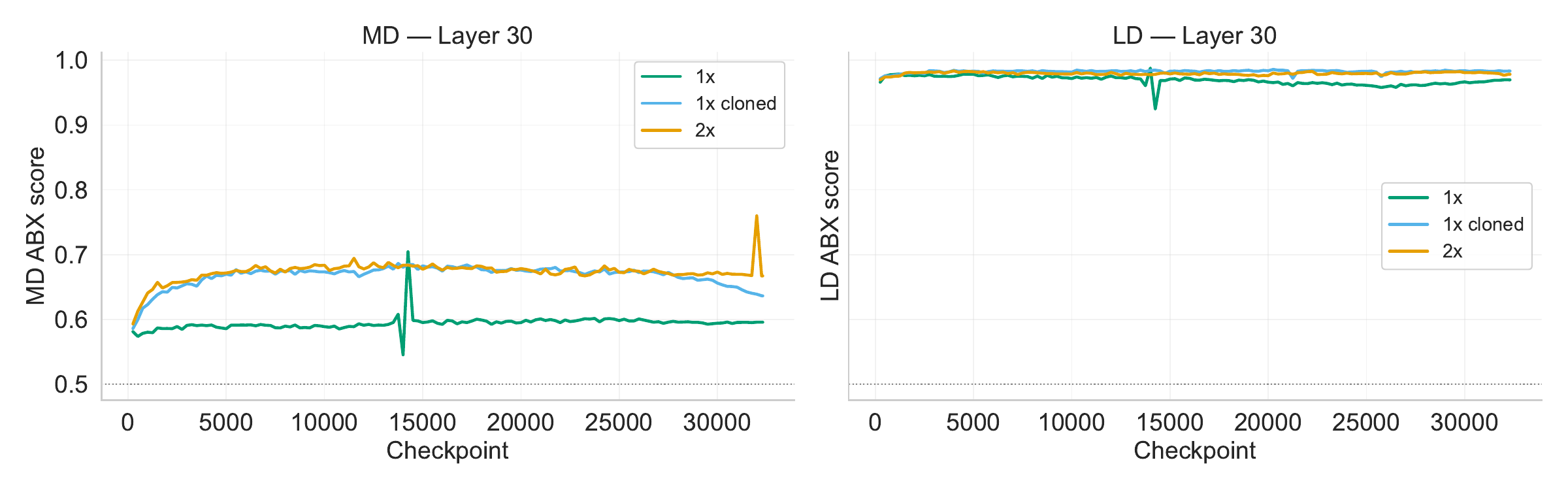}
    \caption{Ablation: A closer look into why the ABX scores for the \SETUP{1× cloned} model drop during training.}
    \label{fig:abx_layerwise_dynamics}
\end{figure}

\paragraph{Training dynamics across checkpoints.}
The drop that we observed in MD for the \SETUP{1× cloned} model in Section~\ref{sec:abx} develops gradually during training and is mainly concentrated in the middle layers. Figure~\ref{fig:abx_layerwise_dynamics} shows there is a breaking point between layers 12--13, while the final layer is stable across models including the \SETUP{1× cloned} model. This suggests that the instability mainly affects the \emph{intermediate} representations rather than the final output layer, which explains why we do not observe a strong difference between the \SETUP{1× cloned} and the \SETUP{2×} model in other evaluations. 

\begin{table*}[t]
\centering
\scriptsize
\setlength{\tabcolsep}{3pt}
\begin{adjustbox}{width=\textwidth}
\begin{tabular*}{1.2\textwidth}{@{\extracolsep{\fill}}lrrrrrrrrrrrrrrrrrr@{}}
\toprule
& \multicolumn{6}{c}{Multilingual (eng×lang)} & \multicolumn{6}{c}{Monolingual (eng×lang)} & \multicolumn{6}{c}{Multilingual (all pairs)} \\
\cmidrule(lr){2-7} \cmidrule(lr){8-13} \cmidrule(lr){14-19}
& \multicolumn{3}{c}{MD} & \multicolumn{3}{c}{LD} & \multicolumn{3}{c}{MD} & \multicolumn{3}{c}{LD} & \multicolumn{3}{c}{MD} & \multicolumn{3}{c}{LD} \\
\cmidrule(lr){2-4} \cmidrule(lr){5-7} \cmidrule(lr){8-10} \cmidrule(lr){11-13} \cmidrule(lr){14-16} \cmidrule(lr){17-19}
 & \SETUP{1×} & \SETUP{2×} & \SETUP{1× cloned} & \SETUP{1×} & \SETUP{2×} & \SETUP{1× cloned} & \SETUP{1×} & \SETUP{2×} & \SETUP{1× cloned} & \SETUP{1×} & \SETUP{2×} & \SETUP{1× cloned} & \SETUP{1×} & \SETUP{2×} & \SETUP{1× cloned} & \SETUP{1×} & \SETUP{2×} & \SETUP{1× cloned} \\
\midrule
fao & 0.645 & 0.639 & 0.649 & 0.803 & 0.770 & 0.755 & 0.629 & 0.644 & 0.710 & 0.791 & 0.772 & 0.792 & 0.773 & 0.789 & 0.773 & 0.790 & 0.778 & 0.764 \\
swe & 0.833 & 0.839 & 0.809 & 0.771 & 0.720 & 0.717 & 0.901 & 0.893 & 0.917 & 0.764 & 0.693 & 0.693 & 0.809 & 0.804 & 0.785 & 0.780 & 0.758 & 0.761 \\
isl & 0.618 & 0.649 & 0.647 & 0.788 & 0.751 & 0.734 & 0.710 & 0.704 & 0.748 & 0.780 & 0.741 & 0.752 & 0.784 & 0.798 & 0.781 & 0.806 & 0.777 & 0.767 \\
fas & 0.742 & 0.743 & 0.738 & 0.815 & 0.770 & 0.787 & 0.812 & 0.791 & 0.824 & 0.821 & 0.730 & 0.736 & 0.750 & 0.765 & 0.748 & 0.820 & 0.779 & 0.780 \\
ekk & 0.866 & 0.812 & 0.776 & 0.756 & 0.747 & 0.779 & 0.902 & 0.870 & 0.891 & 0.769 & 0.738 & 0.775 & 0.788 & 0.796 & 0.777 & 0.801 & 0.777 & 0.772 \\
fin & 0.677 & 0.702 & 0.679 & 0.782 & 0.744 & 0.736 & 0.741 & 0.734 & 0.772 & 0.775 & 0.702 & 0.692 & 0.787 & 0.804 & 0.785 & 0.795 & 0.767 & 0.759 \\
sme & 0.560 & 0.568 & 0.580 & 0.835 & 0.808 & 0.801 & 0.559 & 0.565 & 0.582 & 0.853 & 0.841 & 0.890 & 0.745 & 0.756 & 0.743 & 0.828 & 0.815 & 0.805 \\
ckb & 0.567 & 0.541 & 0.543 & 0.867 & 0.891 & 0.875 & 0.552 & 0.549 & 0.610 & 0.884 & 0.922 & 0.952 & 0.648 & 0.622 & 0.612 & 0.854 & 0.861 & 0.836 \\
pbt & 0.504 & 0.524 & 0.535 & 0.858 & 0.842 & 0.857 & 0.512 & 0.523 & 0.529 & 0.861 & 0.841 & 0.847 & 0.655 & 0.669 & 0.652 & 0.833 & 0.818 & 0.813 \\
sna & 0.580 & 0.595 & 0.602 & 0.840 & 0.793 & 0.788 & 0.602 & 0.613 & 0.636 & 0.810 & 0.780 & 0.827 & 0.749 & 0.770 & 0.751 & 0.828 & 0.809 & 0.795 \\
swh & 0.626 & 0.671 & 0.671 & 0.807 & 0.760 & 0.734 & 0.713 & 0.759 & 0.785 & 0.772 & 0.738 & 0.778 & 0.783 & 0.808 & 0.773 & 0.811 & 0.778 & 0.759 \\
zul & 0.553 & 0.572 & 0.580 & 0.843 & 0.802 & 0.795 & 0.570 & 0.588 & 0.628 & 0.803 & 0.792 & 0.837 & 0.724 & 0.751 & 0.732 & 0.853 & 0.836 & 0.815 \\
\bottomrule
\end{tabular*}
\end{adjustbox}
\caption{ABX scores (last checkpoint, avg over layers).}
\label{tab:abx_per_language}
\end{table*}

\paragraph{Per-language results (Table~\ref{tab:abx_per_language}).}
 
There is a notable difference between mono/bilingual and multilingual models. For mono/bilingual models, the \SETUP{1× cloned} variant often achieves the highest MD scores, while for multilingual models, the \SETUP{2×} variant tends to score highest. Combined with the checkpoint dynamics above, this may be because mono/bilingual models are trained on less data and thus stop training before the MD decline that affects the multilingual \SETUP{1× cloned} model sets in.
 
Results also vary considerably by language. Higher-resource languages like Swedish and Estonian show substantially higher MD scores (0.8--0.9), indicating stronger cross-lingual alignment. Lower-resource languages like Central Kurdish, Southern Pashto, and Northern S\'{a}mi have much lower MD scores (0.5--0.6) paired with high LD scores (0.85--0.95). This suggests that when the model has limited data for a language, it isolates that language in its representation space rather than aligning it with others.

\section{Data Matched Comparison Statistics}
\label{appendix:significance_tests}

We show pairwise score differences between data-matched models with 95\% confidence intervals from paired bootstrap resampling for sentence-level and paragraph level IP in Tables~\ref{tab:sentence_ip_stats} and \ref{tab:paragraph_ip_stats} respectively, as described in Section~\ref{secsec:models}. Pairs are grouped by single"-target and multilingual models. We find that most differences between models are statistically significant in this setup, supporting our results.

\begin{table*}[t]
\small
\begin{tabular*}{0.9\linewidth}{llrrr}
\toprule
Model Family & Language & 1x cloned -- 1x & 2x -- 1x & 2x -- 1x cloned \\
\midrule\addlinespace[2.5pt]
\multirow{12}{*}{Single"-target} & ckb & \textbf{0.048 [0.046, 0.05]} & \textbf{0.058 [0.056, 0.06]} & \textbf{0.01 [0.008, 0.011]} \\
 & ekk & \textbf{0.041 [0.038, 0.045]} & \textbf{0.06 [0.057, 0.064]} & \textbf{0.019 [0.016, 0.022]} \\
 & fao & \textbf{-0.033 [-0.035, -0.03]} & \textbf{0.037 [0.034, 0.04]} & \textbf{0.07 [0.066, 0.073]} \\
 & fas & \textbf{0.009 [0.006, 0.012]} & \textbf{0.095 [0.091, 0.099]} & \textbf{0.086 [0.081, 0.091]} \\
 & fin & \textbf{0.085 [0.081, 0.089]} & \textbf{0.024 [0.021, 0.028]} & \textbf{-0.061 [-0.065, -0.057]} \\
 & isl & \textbf{0.006 [0.002, 0.009]} & \textbf{0.036 [0.032, 0.04]} & \textbf{0.03 [0.027, 0.033]} \\
 & pbt & \textbf{0.009 [0.008, 0.01]} & \textbf{0.012 [0.01, 0.013]} & \textbf{0.003 [0.001, 0.004]} \\
 & sme & -0.001 [-0.003, 0.001] & \textbf{0.03 [0.028, 0.031]} & \textbf{0.031 [0.029, 0.033]} \\
 & sna & \textbf{0.004 [0.002, 0.006]} & \textbf{0.027 [0.025, 0.029]} & \textbf{0.024 [0.021, 0.026]} \\
 & swe & \textbf{0.114 [0.11, 0.118]} & \textbf{0.162 [0.157, 0.167]} & \textbf{0.048 [0.045, 0.052]} \\
 & swh & \textbf{0.021 [0.019, 0.024]} & \textbf{0.007 [0.004, 0.011]} & \textbf{-0.014 [-0.017, -0.011]} \\
 & zul & \textbf{-0.017 [-0.02, -0.015]} & \textbf{0.006 [0.004, 0.008]} & \textbf{0.023 [0.021, 0.026]} \\
\midrule
\multirow{12}{*}{Multilingual} & ckb & -0.001 [-0.003, 0.001] & \textbf{0.011 [0.009, 0.012]} & \textbf{0.011 [0.01, 0.013]} \\
 & ekk & -0.004 [-0.009, 0.001] & \textbf{-0.025 [-0.03, -0.02]} & \textbf{-0.021 [-0.024, -0.017]} \\
 & fao & \textbf{0.033 [0.029, 0.037]} & \textbf{0.021 [0.016, 0.026]} & \textbf{-0.012 [-0.015, -0.009]} \\
 & fas & \textbf{0.012 [0.008, 0.015]} & 0.002 [-0.002, 0.005] & \textbf{-0.01 [-0.012, -0.007]} \\
 & fin & -0.0 [-0.005, 0.005] & \textbf{-0.019 [-0.024, -0.014]} & \textbf{-0.019 [-0.022, -0.016]} \\
 & isl & 0.001 [-0.004, 0.005] & \textbf{-0.013 [-0.018, -0.009]} & \textbf{-0.014 [-0.017, -0.011]} \\
 & pbt & \textbf{0.005 [0.003, 0.006]} & \textbf{0.011 [0.01, 0.013]} & \textbf{0.007 [0.005, 0.008]} \\
 & sme & \textbf{0.043 [0.04, 0.047]} & \textbf{0.036 [0.033, 0.04]} & \textbf{-0.007 [-0.01, -0.005]} \\
 & sna & \textbf{0.025 [0.021, 0.029]} & \textbf{0.015 [0.011, 0.018]} & \textbf{-0.01 [-0.013, -0.007]} \\
 & swe & \textbf{-0.01 [-0.015, -0.005]} & \textbf{-0.031 [-0.037, -0.026]} & \textbf{-0.022 [-0.025, -0.019]} \\
 & swh & \textbf{0.014 [0.01, 0.019]} & -0.004 [-0.008, 0.0] & \textbf{-0.018 [-0.021, -0.015]} \\
 & zul & \textbf{0.041 [0.037, 0.045]} & \textbf{0.029 [0.025, 0.033]} & \textbf{-0.012 [-0.015, -0.009]} \\
\bottomrule
\end{tabular*}
\caption{Differences in sentence-level information parity and 95\% confidence intervals for each pairwise comparison for each language, separated by single"-target and multilingual models. Bold values indicate statistically significant differences ($p < 0.05$) after applying Benjamini-Hochberg correction for each model pair.}\label{tab:sentence_ip_stats}
\end{table*}

\begin{table*}[t]
\small
\begin{tabular*}{0.9\linewidth}{llrrr}
\toprule
Model Family & Language & 1x cloned -- 1x & 2x -- 1x & 2x -- 1x cloned \\
\midrule\addlinespace[2.5pt]
\multirow{12}{*}{Single"-target} & ckb & \textbf{0.018 [0.015, 0.02]} & \textbf{0.021 [0.019, 0.024]} & \textbf{0.004 [0.002, 0.006]} \\
 & ekk & \textbf{0.067 [0.061, 0.073]} & \textbf{0.076 [0.071, 0.082]} & \textbf{0.009 [0.004, 0.014]} \\
 & fao & \textbf{-0.01 [-0.012, -0.007]} & \textbf{0.029 [0.026, 0.032]} & \textbf{0.039 [0.035, 0.042]} \\
 & fas & \textbf{0.064 [0.06, 0.069]} & \textbf{0.099 [0.093, 0.104]} & \textbf{0.034 [0.03, 0.039]} \\
 & fin & \textbf{0.089 [0.083, 0.094]} & \textbf{0.076 [0.07, 0.081]} & \textbf{-0.013 [-0.017, -0.009]} \\
 & isl & \textbf{0.038 [0.032, 0.043]} & \textbf{0.06 [0.054, 0.066]} & \textbf{0.022 [0.018, 0.027]} \\
 & pbt & \textbf{-0.009 [-0.011, -0.006]} & \textbf{0.017 [0.015, 0.018]} & \textbf{0.025 [0.022, 0.028]} \\
 & sme & \textbf{-0.015 [-0.017, -0.013]} & \textbf{0.015 [0.013, 0.016]} & \textbf{0.03 [0.028, 0.032]} \\
 & sna & -0.002 [-0.004, 0.0] & \textbf{0.024 [0.021, 0.026]} & \textbf{0.026 [0.023, 0.028]} \\
 & swe & \textbf{0.092 [0.087, 0.097]} & \textbf{0.111 [0.105, 0.116]} & \textbf{0.018 [0.014, 0.023]} \\
 & swh & \textbf{0.027 [0.024, 0.03]} & \textbf{0.034 [0.03, 0.038]} & \textbf{0.007 [0.003, 0.011]} \\
 & zul & \textbf{-0.012 [-0.015, -0.009]} & \textbf{0.013 [0.011, 0.015]} & \textbf{0.025 [0.022, 0.028]} \\
\midrule
\multirow{12}{*}{Multilingual} & ckb & \textbf{0.08 [0.076, 0.084]} & \textbf{0.087 [0.084, 0.091]} & \textbf{0.008 [0.005, 0.011]} \\
 & ekk & \textbf{0.065 [0.059, 0.07]} & \textbf{0.059 [0.053, 0.065]} & -0.005 [-0.01, -0.001] \\
 & fao & \textbf{0.072 [0.067, 0.077]} & \textbf{0.07 [0.065, 0.075]} & -0.002 [-0.006, 0.002] \\
 & fas & \textbf{0.062 [0.058, 0.066]} & \textbf{0.062 [0.058, 0.067]} & 0.0 [-0.003, 0.004] \\
 & fin & \textbf{0.064 [0.059, 0.07]} & \textbf{0.063 [0.057, 0.069]} & -0.002 [-0.006, 0.003] \\
 & isl & \textbf{0.048 [0.043, 0.053]} & \textbf{0.052 [0.047, 0.057]} & 0.004 [-0.001, 0.008] \\
 & pbt & \textbf{0.064 [0.061, 0.068]} & \textbf{0.072 [0.068, 0.075]} & \textbf{0.008 [0.005, 0.01]} \\
 & sme & \textbf{0.067 [0.063, 0.07]} & \textbf{0.067 [0.063, 0.071]} & 0.0 [-0.003, 0.004] \\
 & sna & \textbf{0.057 [0.054, 0.061]} & \textbf{0.056 [0.052, 0.06]} & -0.001 [-0.004, 0.002] \\
 & swe & \textbf{0.052 [0.046, 0.057]} & \textbf{0.051 [0.045, 0.057]} & -0.001 [-0.005, 0.003] \\
 & swh & \textbf{0.049 [0.044, 0.053]} & \textbf{0.048 [0.043, 0.053]} & -0.001 [-0.004, 0.003] \\
 & zul & \textbf{0.075 [0.07, 0.08]} & \textbf{0.072 [0.067, 0.077]} & -0.003 [-0.007, 0.001] \\
\bottomrule
\end{tabular*}
\caption{Differences in paragraph-level information parity and 95\% confidence intervals for each pairwise comparison for each language, separated by single"-target and multilingual models. Bold values indicate statistically significant differences ($p < 0.05$) after applying Benjamini-Hochberg correction for each model pair.}\label{tab:paragraph_ip_stats}
\end{table*}

\end{document}